\documentclass[letterpaper]{article} 
\usepackage{aaai2027}  
\usepackage[hyphens]{url}  
\usepackage{graphicx} 
\usepackage{natbib} 
\usepackage{caption} 
\usepackage{booktabs}
\usepackage{amsmath}
\usepackage{amssymb}
\usepackage{tikz}
\usetikzlibrary{arrows.meta,positioning,calc}

\definecolor{rawred}{RGB}{196,78,82}
\definecolor{hierorange}{RGB}{221,132,82}
\definecolor{groundblue}{RGB}{76,114,176}
\definecolor{lightgray}{RGB}{244,244,244}
\definecolor{lightblue}{RGB}{232,239,249}
\definecolor{lightred}{RGB}{251,235,234}
\definecolor{lightorange}{RGB}{252,240,229}

\newcommand{\tga}{taxonomy-grounded agent}

\title{Same Feedback, Different Answer: Measuring Run-to-Run Instability in Frontier-Model Customer Feedback Analysis}
\author{Viraj Bagal, Raviraja Ganta, Prabhath Chellingi}
\affiliations{}
\nocopyright

\begin{document}

\maketitle

\begin{abstract}
AI agents are increasingly being programmed to automate knowledge work over large collections of unstructured data.
Such automation requires repeatability: when the underlying evidence is unchanged, the agent's categories, priorities, and counts should not shift materially between runs, even if each individual answer appears plausible.
We introduce a repeat-run evaluation framework that aligns semantically equivalent categories and focuses on two operating metrics: theme churn, the normalized change in the returned category set, and volume disagreement, the change in counts for categories that persist.
We evaluate three recurring customer-feedback tasks across eight frontier models, corpus sizes from 100 to 5,000 records, multiple prompts, and three execution designs: raw generation, taxonomy-free hierarchical decomposition, and a taxonomy-grounded agent (TGA) using persistent themes, subthemes, and record-level predictions.
With Claude Opus 4.8 and the 1,000-record corpus fixed, TGA reduces theme churn by 86--88\% relative to both raw generation and hierarchical decomposition, while matched-theme volumes have zero disagreement.
The taxonomy-grounded agent is more stable than every raw model in the screen, remains more stable at each corpus size, and keeps this advantage when theme matching is made stricter or looser.
Although evaluated on customer feedback, the framework targets repeated synthesis of unstructured corpora more broadly, including financial reports, legal documents, incident records, and scientific literature.
Overall, these results show that taxonomy grounding produces more consistent and repeatable outputs for recurring knowledge work.
\end{abstract}

\section{Introduction}

A customer-operations analyst may ask, ``What are the top customer pain points this month?'' A fluent answer is useful once. An operating answer must also survive repetition: the same records, question, and system configuration should not yield a substantially different theme set or different counts on the next run. This distinction matters because theme identities become dashboard categories, rankings guide roadmaps, and volumes trigger prioritization. Broad language-model evaluations increasingly treat reliability as multidimensional rather than reducible to task accuracy alone \citep{liang2023helm}. Yet repeatability for open-ended corpus analysis remains weakly specified.

Prior work demonstrates sensitivity to prompt wording and formatting \citep{sclar2024quantifying,chatterjee2024posix,errica2025wrong}, self-inconsistency under ambiguity \citep{bartsch2023self,sedova2024know}, and the use of repeated samples to estimate confidence or hallucination risk \citep{lyu2025calibrating,manakul2023selfcheckgpt,farquhar2024detecting}. Customer-feedback analysis creates a different measurement problem. Each answer is an ordered set of natural-language themes, each record may support multiple themes, theme labels can be paraphrased, and counts are comparable only after themes are aligned. Exact-string agreement therefore overstates instability, while a single scalar similarity obscures whether the model changed the categories, their ranks, their counts, or merely the representative examples.

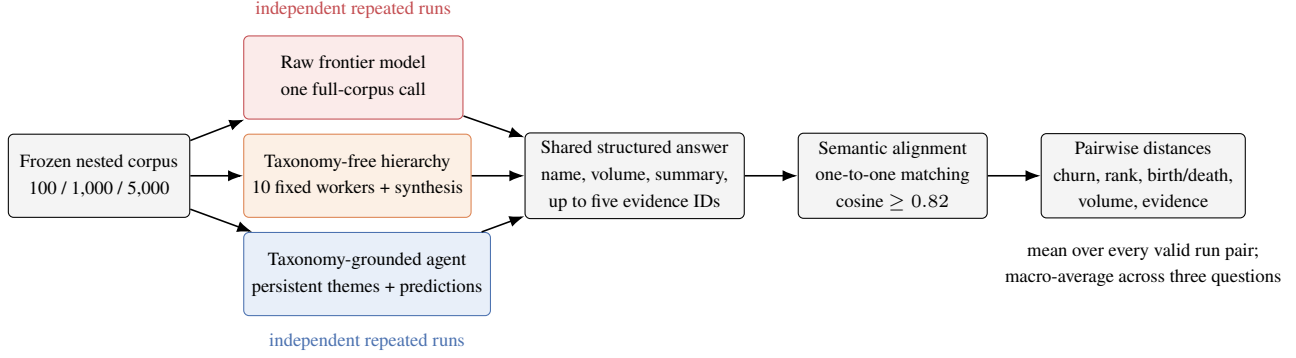
\begin{figure*}[t]
\centering
\begin{tikzpicture}[
  node distance=5mm and 7mm,
  box/.style={draw, rounded corners=2pt, align=center, minimum height=11mm, inner sep=3pt, font=\scriptsize},
  arr/.style={-{Latex[length=2mm]}, semithick},
  note/.style={font=\scriptsize, align=center}
]
\node[box, fill=lightgray, minimum width=24mm] (corpus) {Frozen nested corpus\\100 / 1,000 / 5,000};
\node[box, fill=lightred, draw=rawred, right=7mm of corpus, yshift=13mm, minimum width=29mm] (raw) {Raw frontier model\\one full-corpus call};
\node[box, fill=lightorange, draw=hierorange, right=7mm of corpus, minimum width=29mm] (hier) {Taxonomy-free hierarchy\\10 fixed workers + synthesis};
\node[box, fill=lightblue, draw=groundblue, right=7mm of corpus, yshift=-13mm, minimum width=29mm] (agent) {Taxonomy-grounded agent\\persistent themes + predictions};
\node[box, fill=lightgray, right=7mm of hier, minimum width=29mm] (contract) {Shared structured answer\\name, volume, summary,\\up to five evidence IDs};
\node[box, fill=lightgray, right=7mm of contract, minimum width=25mm] (align) {Semantic alignment\\one-to-one matching\\cosine $\geq 0.82$};
\node[box, fill=lightgray, right=7mm of align, minimum width=27mm] (metrics) {Pairwise distances\\churn, rank, birth/death,\\volume, evidence};
\draw[arr] (corpus) -- (raw);
\draw[arr] (corpus) -- (hier);
\draw[arr] (corpus) -- (agent);
\draw[arr] (raw) -- (contract);
\draw[arr] (hier) -- (contract);
\draw[arr] (agent) -- (contract);
\draw[arr] (contract) -- (align);
\draw[arr] (align) -- (metrics);
\node[note, above=1mm of raw, text=rawred] {independent repeated runs};
\node[note, below=1mm of agent, text=groundblue] {independent repeated runs};
\node[note, below=2mm of metrics] {mean over every valid run pair;\\macro-average across three questions};
\end{tikzpicture}
\caption{Evaluation pipeline. The systems share timestamp windows, questions, output schema, alignment, and metrics. The Opus-controlled experiment distinguishes one-call generation, fixed hierarchical decomposition without taxonomy, and persistent taxonomy grounding.}
\label{fig:pipeline}
\end{figure*}

We study three execution designs. A \emph{raw} frontier model receives every feedback summary in a frozen time window and reclusters the corpus in one call. A \emph{taxonomy-free hierarchical workflow} splits the same records into fixed batches, analyzes them independently, and synthesizes their outputs. A production \emph{\tga} (TGA) instead queries persistent themes, subthemes, and record-level taxonomy predictions before composing the answer. The grounded system does not see a hand-built answer key; category identity and membership are materialized before the question is asked. All systems return the same strict structured-output contract, enabling identical downstream analysis. Figure~\ref{fig:pipeline} summarizes the comparison.

Our contributions are:
\begin{itemize}
    \item A stability-specific evaluation protocol for repeated open-ended, multi-theme corpus analysis. It aligns semantic theme names one-to-one and decomposes instability into theme churn, rank, volume, and representative-evidence distances, with birth/death retained as a directional diagnostic.
    \item A broad screen of eight raw frontier models with ten requested runs for each of three customer-feedback questions, followed by a scale study on nested 100-, 1,000-, and 5,000-record corpora.
    \item A controlled 1,000-record ablation that fixes Claude Opus 4.8, compares three prompt variants, and separates a raw call, a taxonomy-free ten-worker hierarchy, and persistent taxonomy grounding.
    \item Evidence that the most stable eligible raw model still changes much of its operating theme set between runs. With model and corpus controlled, grounded theme churn is 0.104 versus 0.765 for raw generation and 0.841 for the taxonomy-free hierarchy; grounded matched-theme volume disagreement is zero.
    \item Run-level uncertainty estimates, coverage-aware interpretation of conditional metrics, full invalid-output accounting, cost reporting, and a semantic-alignment threshold sweep.
\end{itemize}

This paper measures stability, not correctness. A system can repeat the same wrong answer. We make no claim that the more stable answers are more accurate, useful, or actionable; those properties require separate human-judgment and task-quality studies.

\section{Related Work}

\subsection{Generation Consistency and Prompt Sensitivity}

Recent work shows that repeated LLM evaluations should not ignore nondeterminism: outputs can vary across repeated requests under default sampling, greedy decoding, and hosted settings expected to be deterministic \citep{ouyang2025nondeterminism,song2025good,atil2025nondeterminism}. Self-consistency is often used as a technique for improving reasoning: sample multiple chains and select the modal answer \citep{wang2023selfconsistency}. Other work treats agreement among samples as evidence about confidence. \citet{lyu2025calibrating} compare consistency-based confidence measures across models and reasoning tasks; semantic uncertainty groups generations by meaning before measuring entropy \citep{kuhn2023semantic,farquhar2024detecting}; and black-box confidence-elicitation studies combine prompting, sampling, and aggregation \citep{xiong2024uncertainty}. SelfCheckGPT similarly detects unsupported content through inconsistency across sampled passages \citep{manakul2023selfcheckgpt}. These methods usually reduce repeated generations to a confidence or selection signal. We instead retain disagreement as the object of study and decompose where an operating answer changes.

Prompt-sensitivity research asks whether intent-preserving surface changes alter predictions. Spurious formatting can produce large performance swings \citep{sclar2024quantifying}; POSIX measures likelihood changes under prompt perturbations \citep{chatterjee2024posix}; and sensitivity/consistency metrics expose classification failures across prompt rephrasings \citep{errica2025wrong}. Ambiguity studies also show that a model can remain partly consistent while assigning meaningful probability to alternatives \citep{bartsch2023self,sedova2024know}. Agent benchmarks such as $\tau$-bench evaluate repeated trials through reliability metrics such as pass$^k$ \citep{yao2024taubench}; those tasks have verifiable goal states, while recurring corpus synthesis often has no single gold operating answer. Our repeat-query condition is stricter than prompt perturbation: within an experiment cell, the prompt, corpus, provider configuration, and system path are fixed. The measured variation is run-to-run semantic variation in a structured set answer.

\subsection{Topic Discovery and Customer Feedback}

Classical topic models represent documents as mixtures over latent topics \citep{blei2003lda}; neural topic methods such as BERTopic cluster dense document representations and derive interpretable labels \citep{grootendorst2022bertopic}. TopicGPT instead prompts a language model to generate and assign natural-language topics \citep{pham2024topicgpt}. Topic-model stability has long been recognized as a model-selection concern: independently fitted solutions may yield different topics, motivating explicit stability analysis \citep{greene2014stability}. Our setting differs because the output is an ordered, question-conditioned set of themes rather than a complete corpus model, but it inherits the same identity problem: labels must be aligned before variation can be quantified.

Customer-feedback systems often impose structure to turn free text into operational categories. InsightNet, for example, combines a multi-level taxonomy with multi-label topic extraction for customer reviews \citep{mukku2023insightnet}. Large-scale multi-label classification research studies hierarchical and evolving label spaces \citep{chalkidis2020multilabel,tabatabaei2025hierarchical}, while recent analysis finds that autoregressive language models exhibit distinctive multi-label generation behavior \citep{ma2025multilabel}. These works primarily evaluate predictive quality. We ask whether a question-conditioned aggregation of those labels remains stable across repeated executions.

\subsection{Persistent Context and Semantic Matching}

Retrieval-augmented generation conditions answers on retrieved external evidence \citep{lewis2020rag}; graph-based retrieval and community summaries extend this idea to global corpus questions \citep{edge2024graphrag}. Our grounded arm uses a related system principle---persist domain structure outside the generation---but the experiment is not a retrieval-quality benchmark. It tests whether persistent taxonomy predictions provide a repeatable operating frame relative to reclustering raw records on every call.

Comparing natural-language labels requires semantic rather than lexical equivalence. Sentence-level embeddings make paraphrase-aware similarity practical \citep{reimers2019sbert}. We freeze one embedding model, threshold, and one-to-one assignment rule, then rerun analysis across a threshold range. This prevents an answer with one broad theme from matching several narrower themes in the other run.

\section{Evaluation Framework}

\subsection{Repeated Structured Answers}

An experiment cell fixes a system $s$, corpus $D$, and question $q$. Independent executions produce runs $R_1,\ldots,R_n$. A run is an ordered list of theme results,
\begin{equation}
R_i=[(z_k,v_k,e_k,u_k)]_{k=1}^{K_i},
\end{equation}
where $z_k$ is a theme name, $v_k\geq0$ its reported volume, $e_k$ a summary, and $u_k$ a set of at most five supporting record identifiers. The schema forbids additional fields. Quotes are explicitly excluded.

A feedback record may belong to several themes. Consequently, volumes need not sum to $|D|$, and the supporting-ID sets are representative evidence rather than exhaustive membership lists. The five-ID bound keeps raw and grounded outputs feasible at large scale. It also makes evidence churn a sample-selection diagnostic, not a direct estimate of complete membership churn.

Generation, alignment, and analysis are separate stages. Generation receipts identify the corpus, question, candidate configuration, individual attempts, provider usage, cost, source commit, and raw structured output. An alignment artifact references its generation artifact and frozen embedding configuration. Analysis references both, so metrics can be added or removed without regenerating answers.

\subsection{Semantic Theme Alignment}

For a pair $(R_i,R_j)$, we embed every theme name $z$ with a frozen text embedding model and compute cosine similarities. Edges below threshold $\gamma$ are removed. Among remaining edges, we first maximize match cardinality and then total similarity, producing a one-to-one assignment $A_{ij}$; this is a lexicographic assignment problem related to maximum-weight bipartite matching \citep{kuhn1955hungarian}. We freeze \texttt{text-embedding-3-large} and $\gamma=0.82$ for primary analysis.

Theme names, rather than summaries, define alignment because the operating category is the object whose persistence we test. Summaries and evidence may change even when a category remains the same. The threshold sweep in Section~\ref{sec:robustness} tests whether the headline ordering depends on this choice.

\subsection{Pairwise Stability Metrics}

Let $K_i=|R_i|$, $K_j=|R_j|$, and $M=|A_{ij}|$. Every reported metric is a distance, so lower is more stable.

\paragraph{Theme churn.} We use the complement of the S\o rensen--Dice overlap \citep{dice1945association}:
\begin{equation}
d_{\mathrm{theme}}(i,j)=1-\frac{2M}{K_i+K_j}.
\end{equation}
It is zero when all themes align and one when none align. Theme churn is the primary metric because it remains defined when rank and volume comparisons cannot be made.

\paragraph{Rank instability.} For aligned themes, let $\tau_b$ be Kendall's rank correlation with tie correction \citep{kendall1938rank,kendall1945ties}. We map it to $[0,1]$:
\begin{equation}
d_{\mathrm{rank}}(i,j)=\frac{1-\tau_b}{2}.
\end{equation}
The metric is undefined if fewer than two themes align. Thus zero rank instability means only that the matched themes preserve relative order; it does not imply identical answers.

\paragraph{Theme birth/death.} Directional unmatched-theme rates are averaged:
\begin{equation}
d_{\mathrm{bd}}(i,j)=\frac{1}{2}\left(\frac{K_i-M}{K_i}+\frac{K_j-M}{K_j}\right).
\end{equation}
When $K_i=K_j$, this equals theme churn. We therefore treat it as a directional diagnostic rather than independent evidence.

\paragraph{Volume disagreement.} Over matched pairs $(a,b)\in A_{ij}$,
\begin{equation}
d_{\mathrm{vol}}(i,j)=\frac{1}{M}\sum_{(a,b)\in A_{ij}}
\frac{|v_a-v_b|}{\max(v_a,v_b,1)}.
\end{equation}
It is undefined when $M=0$. The normalization makes a one-record difference more consequential for small themes while bounding every term by one.

\paragraph{Evidence churn.} Let $U_a$ and $U_b$ be the representative supporting-ID sets. We average their Jaccard distances:
\begin{equation}
d_{\mathrm{ev}}(i,j)=\frac{1}{M}\sum_{(a,b)\in A_{ij}}
\left(1-\frac{|U_a\cap U_b|}{|U_a\cup U_b|}\right).
\end{equation}
This metric asks whether the system cites the same examples for a matched theme. It is deliberately secondary because each output contains at most five IDs.

\subsection{Aggregation and Uncertainty}

For $n$ valid runs, we compute all $\binom{n}{2}$ unordered run pairs. The question-level score is the mean pairwise distance,
\begin{equation}
\bar d_{s,D,q}=\binom{n}{2}^{-1}\sum_{i<j}d(R_i,R_j),
\end{equation}
with defined/undefined pair counts retained in the artifact. A system-tier macro score is the unweighted mean of the three question means. This prevents a question with more valid runs, and therefore more pairs, from dominating the result.

Pairwise observations are dependent because each run appears in several pairs. For the scale experiment, we estimate 95\% intervals by deleting one complete run at a time within each question, recomputing the U-statistic, and applying the jackknife standard-error estimator \citep{efron1981jackknife}. We combine question-level variances for the macro mean using a Welch--Satterthwaite $t$ interval. The controlled Opus ablation instead reports per-question run-level bootstrap percentile intervals; when resampling repeats an original run, its artificial self-comparison is omitted. Both procedures resample complete runs rather than treating pair distances as independent.

\section{Experimental Setup}

\subsection{Corpus and Questions}

The corpus contains anonymized, multilingual enterprise customer-support feedback. We expose no raw text, record identifiers, tenant identifiers, product names, or user metadata. Records are ordered by timestamp and record identifier to create immutable nested prefixes. Every presented window starts at 2026-04-01 15:16:43 UTC; exclusive endpoints are 2026-06-12 03:26:12 for 100 records, 2026-06-16 16:30:56 for 1,000, and 2026-06-29 15:15:55 for 5,000.

We freeze three operational questions:
\begin{enumerate}
    \item summarize the top five reasons customers provide feedback;
    \item identify the top three customer pain points and affected workflows; and
    \item identify and prioritize the top feature requests, their underlying needs, and frequency.
\end{enumerate}
The full prompts request a volume and concise summary for every theme and explicitly state that customer quotes are unnecessary. The question text does not change between repeated runs.

\subsection{Systems Compared}

\paragraph{Raw frontier model.} The raw arm receives the complete list of feedback summaries, timestamps, and record IDs in the window, plus the question and JSON schema. It has no production taxonomy predictions. Each execution therefore induces its own grouping of the records. Provider adapters use structured-output facilities when available and validate the same schema after generation.

\paragraph{Taxonomy-free hierarchical workflow.} This condition uses fixed rather than adaptive decomposition. It splits the 1,000-record corpus into ten ordered batches of 100, invokes ten parallel structured workers, and passes their ordered outputs to one structured synthesis call. It has no tools, retrieval, planning, adaptive branching, taxonomy, or persistent record predictions. Thus every final answer uses 11 stochastic model calls while holding both the corpus and model fixed.

\paragraph{Taxonomy-grounded agent.} The TGA is a production supervisor/worker architecture with general-purpose workers and workers that query the persistent taxonomy store. It queries persistent themes, subthemes, and record-level predictions constrained to the frozen timestamp window. The agent is prompted to return the shared answer schema and omit quotes; no deterministic post-processing selects themes or evidence. Experiments~1--2 use a Claude Sonnet 4.6 supervisor with provider-default reasoning. Across 105 requested TGA answers in Experiments~1--2, 98 valid outputs are evaluated; retained traces for these outputs contain no subagent events, so the supervisor performs the analysis directly. Experiment~3 explicitly pins the same production architecture to Claude Opus 4.8. Exact record-level retrieval parity is neither expected nor required because the TGA retrieves aggregate taxonomy objects; corpus parity is enforced by auditing the timestamp window in its analytical queries.

Experiments~1--2 compare complete deployed system designs, so model and architecture differ. Experiment~3 controls the language model and adds the taxonomy-free hierarchy to test whether decomposition alone accounts for the stability gap. All arms use provider-default reasoning settings. A sanitized supplemental model-ID table lists the provider-facing identifiers retained in artifacts or resolved from the production model catalog.

\subsection{Experiment 1: Frontier-Model Screen}

At 100 records, we request ten independent runs for every model--question cell. The eight raw configurations span OpenAI, Bedrock Converse, Google GenAI, and Fireworks provider interfaces. Output limits are 32,768 tokens in this screen. Provider defaults are used for sampling because the compared frontier interfaces do not expose a common, semantically equivalent temperature control.

A raw model is eligible for selection only if every question yields at least nine valid structured outputs. We rank eligible models by macro theme churn, then rank instability, then birth/death. Invalid attempts remain in the artifact store and in cost totals; we do not rerun a cell until it happens to pass. The TGA is a comparator, not a candidate in raw-model selection.

\subsection{Experiment 2: Scale Comparison}

We carry the most stable eligible raw configuration into the scale experiment and compare it with the TGA at 100 and 1,000 records, requesting ten fresh runs for each question. The 100-record raw cells are regenerated under the scale execution control (one call at a time) rather than reused from the concurrent model screen. This isolates scale from within-arm concurrency changes.

The 5,000-record tier is a preconfigured operational stress test with five requested runs per question. Each raw call contains roughly 744,000 input tokens. To remove token-throughput throttling as a confounder, calls run serially, provider retries are disabled, the output limit is 8,192 tokens, and each billable request is followed by a recorded 120--150 second cooldown. The experiment harness applies exponential backoff to zero-usage retryable failures. This produced zero throughput throttles, but only three valid raw answers per question, so 5,000-record uncertainty is necessarily wider.

\subsection{Experiment 3: Opus-Controlled Prompt and Architecture Ablation}

We freeze Claude Opus 4.8, provider-default reasoning, the 1,000-record corpus, the three questions, and the shared schema. No condition requests hidden reasoning text in the output. Prompt sensitivity is evaluated first, without inspecting a grounded result: raw and taxonomy-free systems each use three frozen system prompts---\emph{minimal} (task and schema only), \emph{bottom-up} (record-first extraction followed by semantic consolidation), and \emph{consistency-first} (canonical labels, stable boundaries, non-overlap, and deterministic tie-breaking). This is $2$ systems $\times 3$ prompts $\times 3$ questions $\times 10$ runs, or 180 final outputs.

A prompt is eligible only if every one of its six system--question cells has at least 8/10 valid outputs. We select the lowest unweighted macro theme churn, breaking ties by birth/death, volume disagreement, and then rank instability when coverage is adequate. Exact selected raw and workflow cells are reused in the architecture comparison; they are not regenerated after selection. We then add 30 fresh TGA outputs ($3$ questions $\times 10$ runs), yielding 210 planned full answers.

Raw and taxonomy-free model calls use deterministic aliases \texttt{R0000}--\texttt{R0999}, mapped back to immutable corpus IDs. Both use the same evidence normalizer: preserve first-seen order, discard out-of-scope aliases, remove duplicates, and retain at most five IDs and never more than the reported volume. This prevents identifier transport from becoming an uncontrolled difference. Before full execution, all configurations pass two-run smoke tests; the TGA smoke test is rerun after its production runtime changes and is pinned to runtime commit \texttt{1f2634478f7a}.

\subsection{Validity, Cost, and Traceability}

An answer is valid only if it parses into the strict schema, contains no more than five unique evidence IDs per theme, and uses IDs from the frozen corpus where IDs are returned. Metrics use valid outputs only; the paper reports validity alongside stability. Costs include every billable physical generation attempt, including malformed answers and retries. Raw cost uses captured provider token usage and the rate vector frozen in each generation artifact. TGA cost uses the production session estimate over its internal model calls; retained cost artifacts expose aggregate session-level usage, not a supervisor/worker/retrieval split. Embedding and local analysis costs are excluded.

Every stage is immutable and content addressed. Generation receipts include source commit, dependency lock hash, corpus hash, candidate parameters, timestamps, attempts, usage, output, and raw-provider model identifiers where the harness calls providers directly. TGA receipts add session IDs, trace IDs, usage events, and production model selectors. Alignment receipts add the embedding model, threshold, and generation reference. Analysis receipts add metric versions and alignment references. The Opus ablation uses generation source \texttt{55c4159ef6c0} and analysis source \texttt{96060b122abb}; its prompt and architecture exports carry SHA-256 digests and resolve every result to attempts, prompts, model receipts, corpus receipts, and usage. This lineage permits post-hoc metric and threshold analysis without new generation.

\section{Results}

\subsection{Experiment 1: Frontier-Model Variation}

Table~\ref{tab:model-screen} reports the screen. All raw models exhibit substantial theme churn. Claude Fable 5 is the most stable eligible raw model at 0.616, followed by GPT-5.6 Sol at 0.706. Three candidates fail the pre-specified nine-valid-per-question floor. Their metrics remain visible but do not participate in winner selection. The TGA returns 30 valid answers and has zero theme, rank, birth/death, and volume distance across 135 run pairs.

Low conditional rank scores do not contradict high theme churn. Fable's 0.000 rank instability says that themes which align preserve their order; its 0.616 churn says many themes do not align at all. Theme churn is therefore the selection metric, while rank is a diagnostic conditional on theme survival.

\begin{table*}[t]
\centering
\caption{Experiment 1 at 100 records. Valid outputs are shown for feedback reasons / pain points / feature requests. Lower stability metrics are better. Ineligible candidates fail the validity floor; the TGA is a comparator. Cost includes invalid and retried generation attempts.}
\label{tab:model-screen}
\small
\begin{tabular}{lccccccc}
\toprule
Candidate & Valid & Churn & Rank & Birth/death & Volume & Cost (USD) & Eligible \\
\midrule
\textbf{Taxonomy-grounded agent} & 10/10/10 & \textbf{0.000} & \textbf{0.000} & \textbf{0.000} & \textbf{0.000} & 14.488 & Comparator \\
\textbf{Claude Fable 5} & 10/10/9 & \textbf{0.616} & \textbf{0.000} & 0.616 & 0.116 & 8.809 & Yes \\
GLM 5.2 & 9/10/8 & 0.657 & 0.192 & 0.657 & 0.152 & 3.237 & No \\
Claude Opus 4.8 & 9/10/6 & 0.669 & 0.000 & 0.669 & 0.108 & 7.218 & No \\
GPT-5.6 Sol & 10/10/10 & 0.706 & 0.073 & 0.706 & \textbf{0.082} & 4.083 & Yes \\
Kimi K2.7 Code & 9/10/9 & 0.715 & 0.056 & 0.715 & 0.249 & 2.276 & Yes \\
Qwen 3.7 Plus & 5/3/6 & 0.817 & 0.167 & 0.706 & 0.167 & 0.784 & No \\
DeepSeek V4 Pro & 9/10/10 & 0.854 & 0.074 & 0.854 & 0.136 & 2.009 & Yes \\
Gemini 3.1 Pro Preview & 10/10/10 & 0.877 & 0.067 & 0.877 & 0.128 & 5.149 & Yes \\
\bottomrule
\end{tabular}
\end{table*}

\subsection{Experiment 2: Stability Across Corpus Scale}

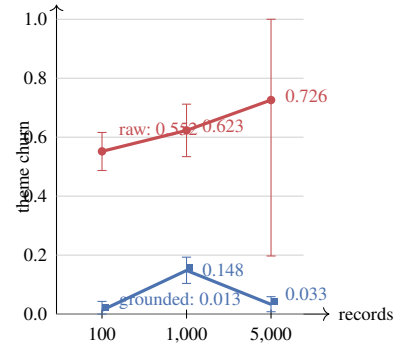
\begin{figure}[t]
\centering
\begin{tikzpicture}[x=0.86cm,y=3.9cm,font=\scriptsize]
  \draw[->] (0,0) -- (4.2,0) node[right] {records};
  \draw[->] (0,0) -- (0,1.05);
  \node[rotate=90] at (-0.52,0.5) {theme churn};
  \foreach \y/\lab in {0/0.0,0.2/0.2,0.4/0.4,0.6/0.6,0.8/0.8,1.0/1.0}{
    \draw[gray!35] (0,\y) -- (3.8,\y);
    \node[left] at (0,\y) {\lab};
  }
  \foreach \x/\lab in {0.7/100,2.0/{1,000},3.3/{5,000}}{
    \draw (\x,0.015) -- (\x,-0.015);
    \node[below] at (\x,-0.015) {\lab};
  }
  \foreach \x/\lo/\hi in {0.7/0.487/0.616,2.0/0.534/0.712,3.3/0.197/1.000}{
    \draw[rawred] (\x,\lo) -- (\x,\hi);
    \draw[rawred] (\x-0.07,\lo) -- (\x+0.07,\lo);
    \draw[rawred] (\x-0.07,\hi) -- (\x+0.07,\hi);
  }
  \foreach \x/\lo/\hi in {0.7/0.000/0.043,2.0/0.104/0.193,3.3/0.008/0.059}{
    \draw[groundblue] (\x,\lo) -- (\x,\hi);
    \draw[groundblue] (\x-0.07,\lo) -- (\x+0.07,\lo);
    \draw[groundblue] (\x-0.07,\hi) -- (\x+0.07,\hi);
  }
  \draw[very thick, rawred] (0.7,0.552) -- (2.0,0.623) -- (3.3,0.726);
  \draw[very thick, groundblue] (0.7,0.013) -- (2.0,0.148) -- (3.3,0.033);
  \foreach \x/\y in {0.7/0.552,2.0/0.623,3.3/0.726}{\fill[rawred] (\x,\y) circle (1.6pt);}
  \foreach \x/\y in {0.7/0.013,2.0/0.148,3.3/0.033}{\fill[groundblue] (\x,\y) rectangle +(2.4pt,2.4pt);}
  \node[rawred, anchor=west] at (0.82,0.63) {raw: 0.552};
  \node[rawred, anchor=west] at (2.10,0.64) {0.623};
  \node[rawred, anchor=west] at (3.38,0.74) {0.726};
  \node[groundblue, anchor=west] at (0.82,0.045) {grounded: 0.013};
  \node[groundblue, anchor=west] at (2.10,0.15) {0.148};
  \node[groundblue, anchor=west] at (3.38,0.07) {0.033};
\end{tikzpicture}
\caption{Macro mean pairwise theme churn with 95\% complete-run jackknife intervals. The 5,000-record raw interval is wide because only three valid runs per question survived.}
\label{fig:scale}
\end{figure}

Figure~\ref{fig:scale} and Table~\ref{tab:scale} show the requested 100-, 1,000-, and 5,000-record sequence. Raw Fable theme churn increases from 0.552 to 0.623 and 0.726. A least-squares descriptive slope over these three points is $+0.0325$ churn per additional 1,000 records. The TGA's corresponding churn values are 0.013, 0.148, and 0.033, substantially below the raw model at every presented tier.

The gap is operationally large. At 100 records the raw model changes more than half of its aligned theme-set mass between two runs on average; at 5,000 it changes nearly three quarters. In contrast, the TGA retains nearly all theme identity in the stress test. The TGA's 1,000-record churn is concentrated in theme appearance/disappearance rather than count movement: volume disagreement remains exactly zero.

\begin{table*}[t]
\centering
\caption{Stability across the presented scale sequence. Coverage in parentheses is the number of defined pairs for a conditional metric over all valid run pairs. The 5,000-record tier requested five rather than ten runs per cell.}
\label{tab:scale}
\small
\begin{tabular}{llccccccc}
\toprule
Records & Arm & Valid & Pairs & Theme churn [95\% CI] & Rank (coverage) & Birth/death & Volume (coverage) & Cost \\
\midrule
100 & Raw Fable & 30/30 & 135 & 0.552 [0.487, 0.616] & 0.023 (91/135) & 0.551 & 0.083 (123/135) & \$8.790 \\
100 & TGA & 29/30 & 126 & \textbf{0.013 [0.000, 0.043]} & 0.026 (126/126) & \textbf{0.013} & \textbf{0.000 (126/126)} & \$14.934 \\
1,000 & Raw Fable & 30/30 & 135 & 0.623 [0.534, 0.712] & 0.015 (72/135) & 0.623 & 0.098 (111/135) & \$51.249 \\
1,000 & TGA & 27/30 & 109 & \textbf{0.148 [0.104, 0.193]} & 0.030 (109/109) & \textbf{0.144} & \textbf{0.000 (109/109)} & \$17.605 \\
5,000 & Raw Fable & 9/15 & 9 & 0.726 [0.197, 1.000] & 0.000 (4/9) & 0.726 & 0.075 (6/9) & \$143.245 \\
5,000 & TGA & 12/15 & 19 & \textbf{0.033 [0.008, 0.059]} & 0.000 (19/19) & \textbf{0.029} & \textbf{0.000 (19/19)} & \$14.812 \\
\bottomrule
\end{tabular}
\end{table*}

\subsection{Experiment 3: Opus-Controlled Ablation}

The prompt-selection pass tests minimal, bottom-up, and consistency-first prompts before inspecting grounded results. Minimal and bottom-up are ineligible because at least one taxonomy-free cell falls below 8/10 valid outputs. Consistency-first is the only eligible prompt and also has the lowest macro theme churn, 0.803. Prompt wording changes the measured values, but none of the raw or taxonomy-free prompt configurations approaches the grounded result.

With Opus 4.8 fixed, the TGA records 0.104 macro theme churn, compared with 0.765 for the raw call and 0.841 for the taxonomy-free hierarchy (Table~\ref{tab:opus-ablation}). These are reductions of 86.4\% and 87.6\% from the displayed values. Decomposition into ten workers and a synthesis call therefore does not explain the grounded system's stability. Its matched-theme volumes are identical in all 117 valid run pairs. Rank is also defined for all 117 pairs, versus 47/135 for raw and 43/118 for taxonomy-free; the apparently low taxonomy-free rank score of 0.047 describes only the small intersection that survives its high churn.

The separation holds for each question. Run-level bootstrap intervals are [0.659, 0.826], [0.600, 0.842], and [0.759, 0.870] for raw Opus, while grounded intervals are [0.000, 0.000], [0.000, 0.281], and [0.078, 0.156]. A degenerate interval means no variation was observed in the ten runs, not that population variance is proven to be zero. Representative-evidence churn does not separate the systems: TGA is 0.461 and raw is 0.448. Because each answer exposes at most five examples, this metric measures sample-evidence turnover rather than complete taxonomy membership.

\begin{table*}[t]
\centering
\caption{Opus-controlled experiment at 1,000 records. Lower metric values are more stable. Raw and taxonomy-free use the selected consistency-first prompt while TGA uses its production prompt; cost covers each 30-output segment. Rank coverage is defined pairs / all valid pairs. Per-question rows give theme churn with 95\% run-level bootstrap percentile intervals.}
\label{tab:opus-ablation}
\scriptsize
\begin{tabular}{lcccccccc}
\multicolumn{9}{l}{\textbf{A. Architecture comparison after prompt selection}} \\
\toprule
System & Valid & Theme & Birth/death & Volume & Rank & Evidence & Rank cov. & Cost \\
\midrule
Raw Opus & 10/10/10 & 0.765 & 0.764 & 0.244 & 0.105 & \textbf{0.448} & 47/135 & \$21.821 \\
Taxonomy-free hierarchy & 10/10/8 & 0.841 & 0.840 & 0.247 & 0.047 & 0.507 & 43/118 & \$37.299 \\
\textbf{Taxonomy-grounded agent} & \textbf{10/9/9} & \textbf{0.104} & \textbf{0.099} & \textbf{0.000} & \textbf{0.008} & 0.461 & \textbf{117/117} & \$16.536 \\
\bottomrule
\end{tabular}

\vspace{1mm}
\begin{tabular}{lccc}
\multicolumn{4}{l}{\textbf{B. Theme churn by question}} \\
\toprule
System & Feedback reasons & Customer pain points & Feature requests \\
\midrule
Raw Opus & 0.756 [0.659, 0.826] & 0.719 [0.600, 0.842] & 0.820 [0.759, 0.870] \\
Taxonomy-free hierarchy & 0.689 [0.595, 0.766] & 0.904 [0.783, 0.984] & 0.931 [0.906, 0.949] \\
\textbf{Taxonomy-grounded agent} & \textbf{0.000 [0.000, 0.000]} & \textbf{0.185 [0.000, 0.281]} & \textbf{0.126 [0.078, 0.156]} \\
\bottomrule
\end{tabular}
\end{table*}

\subsection{Counts, Rank, and Evidence}

Matched-theme volumes provide the clearest operating-answer result. Raw volume disagreement is 0.083, 0.098, and 0.075 at the three tiers; TGA volume disagreement is 0.000 in every defined pair. Thus, once a grounded theme is present in both runs, its reported volume does not move. This directly tests the expectation that semantic judgment may vary while theme counts remain stable.

Rank requires more caution. At 1,000 records, raw rank instability is only 0.015, but it is defined for 72 of 135 pairs; at 5,000 it is zero for only 4 of 9 pairs. The low values describe the relative ordering of a small surviving intersection. The TGA supports rank calculation for every valid pair at all three tiers. Reporting rank without coverage would invert the practical interpretation.

Evidence churn remains nonzero for both systems: raw/TGA values are 0.318/0.207 at 100, 0.295/0.432 at 1,000, and 0.485/0.395 at 5,000. A stable category can rotate among five recent examples, especially as the corpus grows. Evidence churn therefore should not be treated as taxonomy churn.

\subsection{Structured-Output Reliability and Cost}

Stability is conditional on obtaining a valid answer, so validity is a separate outcome. In the 100-record screen, validity ranges from 14/30 for Qwen 3.7 Plus to 30/30 for GPT-5.6 Sol, Gemini 3.1 Pro Preview, and the TGA. In the 100- and 1,000-record scale cells, raw Fable yields 60/60 valid outputs while the TGA yields 56/60. At 5,000 records, Fable yields 9/15 valid outputs despite zero throughput throttles; four failures return the answer field as a string and two return an empty object. The TGA yields 12/15 valid outputs.

Full-corpus raw cost grows from \$8.79 at 100 to \$51.25 at 1,000 and \$143.25 at 5,000. TGA cost is \$14.93, \$17.61, and \$14.81. These totals use system-specific accounting and exclude amortized taxonomy ingestion cost, so they are descriptive rather than a controlled price benchmark. They nevertheless show the operational consequence of repeatedly transmitting the entire corpus versus querying a preprocessed context layer.

The Opus-controlled program requests 210 full answers and yields 196 valid outputs: raw 90/90, taxonomy-free 78/90, and TGA 28/30. Taxonomy-free failures are structured-output violations, not throughput throttles. The two invalid TGA trials are an empty structured object and an exact-window query-audit failure. One additional TGA feature-request attempt returns HTTP 502 with zero usage; its retry succeeds and the failure is recorded as unbilled. Full-generation costs are \$65.77 for raw prompt sensitivity, \$114.78 for taxonomy-free prompt sensitivity, and \$16.54 for fresh TGA runs. Including \$90.72 of preliminary, hardening, and original smoke and \$8.98 of refreshed TGA smoke, the complete program contains 66 submitted generation jobs, 306 physical provider attempts, and costs \$296.79.

\section{Robustness and Diagnostics}
\label{sec:robustness}

\subsection{Alignment-Threshold Sensitivity}

The primary matcher uses cosine threshold 0.82. To test dependence on this choice, we reuse stored theme embeddings and rematerialize alignment and analysis at thresholds $0.76,0.78,\ldots,0.90$. This creates no new model generations or embedding calls. All primary-threshold analysis artifacts reproduce exactly, including means, defined-pair counts, and intervals.

At thresholds 0.76, 0.82, and 0.90, raw Fable churn is 0.357/0.552/0.812 at 100 records, 0.439/0.623/0.814 at 1,000, and 0.667/0.726/0.889 at 5,000. TGA remains lower: 0.000/0.013/0.040, 0.148/0.148/0.148, and 0.033/0.033/0.033. The system ordering never reverses. At 1,000 and 5,000 records, TGA churn is unchanged across thresholds because recurring taxonomy labels align near-verbatim and the remaining unmatched categories stay unmatched throughout the sweep.

\subsection{Uncertainty and Pair Dependence}

At 100 records, the 95\% churn intervals are [0.487, 0.616] for Fable and [0.000, 0.043] for TGA. At 1,000 they are [0.534, 0.712] and [0.104, 0.193]. These complete-run jackknife intervals are separated. The 5,000-record intervals are [0.197, 1.000] and [0.008, 0.059]; the raw interval is broad because deleting one of only three valid runs leaves little pairwise information. We present the point as a stress test, not as a precision-equivalent replacement for the ten-run tiers.

\section{Limitations and Ethical Considerations}

Stability is not quality: repeatability can preserve a systematic error, and this study does not compare themes against a gold taxonomy, operator judgments, or decision outcomes. Prior work similarly distinguishes agreement from correctness \citep{lyu2025calibrating,kuhn2023semantic}. The corpus comes from one private deployment and a narrow period, so customer mix, language distribution, taxonomy maturity, channel, and bursty arrivals may affect clustering.

TGA stability is conditional on a fixed taxonomy and prediction artifact; rebuilding the taxonomy could move instability to ingestion time. Thus the result is that taxonomy grounding amortizes and stabilizes repeated question answering after ingestion, not that construction-time stability is solved. Other limits remain; private feedback, IDs, endpoints, and organization identifiers are not released.

\bibliography{aaai2027}

\begin{thebibliography}{32}
\providecommand{\natexlab}[1]{#1}

\bibitem[{At{\i}l et~al.(2025)At{\i}l, Aykent, Chittams, Fu, Passonneau,
  Radcliffe, Rajan~Rajagopal, Sloan, Tudrej, Ture, Wu, Xu, and
  Baldwin}]{atil2025nondeterminism}
At{\i}l, B.; Aykent, S.; Chittams, A.; Fu, L.; Passonneau, R.~J.; Radcliffe,
  E.; Rajan~Rajagopal, G.; Sloan, A.; Tudrej, T.; Ture, F.; Wu, Z.; Xu, L.; and
  Baldwin, B. 2025.
\newblock Non-Determinism of ``Deterministic'' {LLM} System Settings in Hosted
  Environments.
\newblock In \emph{Proceedings of the 5th Workshop on Evaluation and Comparison
  of NLP Systems}, 135--148. Association for Computational Linguistics.

\bibitem[{Bartsch et~al.(2023)Bartsch, Jorgensen, Rosati, Hoelscher-Obermaier,
  and Pfau}]{bartsch2023self}
Bartsch, H.; Jorgensen, O.; Rosati, D.; Hoelscher-Obermaier, J.; and Pfau, J.
  2023.
\newblock Self-Consistency of Large Language Models under Ambiguity.
\newblock In \emph{Proceedings of the 6th BlackboxNLP Workshop: Analyzing and
  Interpreting Neural Networks for NLP}, 89--105. Association for Computational
  Linguistics.

\bibitem[{Blei, Ng, and Jordan(2003)}]{blei2003lda}
Blei, D.~M.; Ng, A.~Y.; and Jordan, M.~I. 2003.
\newblock Latent Dirichlet Allocation.
\newblock \emph{Journal of Machine Learning Research}, 3: 993--1022.

\bibitem[{Chalkidis et~al.(2020)Chalkidis, Fergadiotis, Kotitsas, Malakasiotis,
  Aletras, and Androutsopoulos}]{chalkidis2020multilabel}
Chalkidis, I.; Fergadiotis, M.; Kotitsas, S.; Malakasiotis, P.; Aletras, N.;
  and Androutsopoulos, I. 2020.
\newblock An Empirical Study on Large-Scale Multi-Label Text Classification
  Including Few and Zero-Shot Labels.
\newblock In \emph{Proceedings of the 2020 Conference on Empirical Methods in
  Natural Language Processing}, 7503--7515. Association for Computational
  Linguistics.

\bibitem[{Chatterjee et~al.(2024)Chatterjee, Renduchintala, Bhatia, and
  Chakraborty}]{chatterjee2024posix}
Chatterjee, A.; Renduchintala, H. S. V. N. S.~K.; Bhatia, S.; and Chakraborty,
  T. 2024.
\newblock {POSIX}: A Prompt Sensitivity Index for Large Language Models.
\newblock In \emph{Findings of the Association for Computational Linguistics:
  EMNLP 2024}, 14550--14565. Association for Computational Linguistics.

\bibitem[{Dice(1945)}]{dice1945association}
Dice, L.~R. 1945.
\newblock Measures of the Amount of Ecologic Association Between Species.
\newblock \emph{Ecology}, 26(3): 297--302.

\bibitem[{Edge et~al.(2024)Edge, Trinh, Cheng, Bradley, Chao, Mody, Truitt, and
  Larson}]{edge2024graphrag}
Edge, D.; Trinh, H.; Cheng, N.; Bradley, J.; Chao, A.; Mody, A.; Truitt, S.;
  and Larson, J. 2024.
\newblock From Local to Global: A Graph {RAG} Approach to Query-Focused
  Summarization.
\newblock \emph{arXiv preprint arXiv:2404.16130}.

\bibitem[{Efron(1981)}]{efron1981jackknife}
Efron, B. 1981.
\newblock Nonparametric Estimates of Standard Error: The Jackknife, the
  Bootstrap and Other Methods.
\newblock \emph{Biometrika}, 68(3): 589--599.

\bibitem[{Errica et~al.(2025)Errica, Sanvito, Siracusano, and
  Bifulco}]{errica2025wrong}
Errica, F.; Sanvito, D.; Siracusano, G.; and Bifulco, R. 2025.
\newblock What Did I Do Wrong? Quantifying {LLM}s' Sensitivity and Consistency
  to Prompt Engineering.
\newblock In \emph{Proceedings of the 2025 Conference of the Nations of the
  Americas Chapter of the Association for Computational Linguistics: Human
  Language Technologies}, 1543--1558. Association for Computational
  Linguistics.

\bibitem[{Farquhar et~al.(2024)Farquhar, Kossen, Kuhn, and
  Gal}]{farquhar2024detecting}
Farquhar, S.; Kossen, J.; Kuhn, L.; and Gal, Y. 2024.
\newblock Detecting Hallucinations in Large Language Models Using Semantic
  Entropy.
\newblock \emph{Nature}, 630: 625--630.

\bibitem[{Greene, O'Callaghan, and Cunningham(2014)}]{greene2014stability}
Greene, D.; O'Callaghan, D.; and Cunningham, P. 2014.
\newblock How Many Topics? Stability Analysis for Topic Models.
\newblock In \emph{Machine Learning and Knowledge Discovery in Databases},
  498--513. Springer.

\bibitem[{Grootendorst(2022)}]{grootendorst2022bertopic}
Grootendorst, M. 2022.
\newblock {BERTopic}: Neural Topic Modeling with a Class-Based {TF-IDF}
  Procedure.
\newblock \emph{arXiv preprint arXiv:2203.05794}.

\bibitem[{Kendall(1938)}]{kendall1938rank}
Kendall, M.~G. 1938.
\newblock A New Measure of Rank Correlation.
\newblock \emph{Biometrika}, 30(1/2): 81--93.

\bibitem[{Kendall(1945)}]{kendall1945ties}
Kendall, M.~G. 1945.
\newblock The Treatment of Ties in Ranking Problems.
\newblock \emph{Biometrika}, 33(3): 239--251.

\bibitem[{Kuhn(1955)}]{kuhn1955hungarian}
Kuhn, H.~W. 1955.
\newblock The Hungarian Method for the Assignment Problem.
\newblock \emph{Naval Research Logistics Quarterly}, 2(1--2): 83--97.

\bibitem[{Kuhn, Gal, and Farquhar(2023)}]{kuhn2023semantic}
Kuhn, L.; Gal, Y.; and Farquhar, S. 2023.
\newblock Semantic Uncertainty: Linguistic Invariances for Uncertainty
  Estimation in Natural Language Generation.
\newblock In \emph{International Conference on Learning Representations}.

\bibitem[{Lewis et~al.(2020)Lewis, Perez, Piktus, Petroni, Karpukhin, Goyal,
  K{\"u}ttler, Lewis, Yih, Rockt{\"a}schel, Riedel, and Kiela}]{lewis2020rag}
Lewis, P.; Perez, E.; Piktus, A.; Petroni, F.; Karpukhin, V.; Goyal, N.;
  K{\"u}ttler, H.; Lewis, M.; Yih, W.-t.; Rockt{\"a}schel, T.; Riedel, S.; and
  Kiela, D. 2020.
\newblock Retrieval-Augmented Generation for Knowledge-Intensive {NLP} Tasks.
\newblock In \emph{Advances in Neural Information Processing Systems},
  volume~33, 9459--9474.

\bibitem[{Liang et~al.(2023)Liang, Bommasani, Lee, Tsipras, Soylu, Yasunaga,
  Zhang, Narayanan, Wu, Kumar et~al.}]{liang2023helm}
Liang, P.; Bommasani, R.; Lee, T.; Tsipras, D.; Soylu, D.; Yasunaga, M.; Zhang,
  Y.; Narayanan, D.; Wu, Y.; Kumar, A.; et~al. 2023.
\newblock Holistic Evaluation of Language Models.
\newblock \emph{Transactions on Machine Learning Research}.

\bibitem[{Lyu et~al.(2025)Lyu, Shridhar, Malaviya, Zhang, Elazar, Tandon,
  Apidianaki, Sachan, and Callison-Burch}]{lyu2025calibrating}
Lyu, Q.; Shridhar, K.; Malaviya, C.; Zhang, L.; Elazar, Y.; Tandon, N.;
  Apidianaki, M.; Sachan, M.; and Callison-Burch, C. 2025.
\newblock Calibrating Large Language Models with Sample Consistency.
\newblock \emph{Proceedings of the AAAI Conference on Artificial Intelligence},
  39(18): 19260--19268.

\bibitem[{Ma et~al.(2025)Ma, Chochlakis, Pandiyan, Thomason, and
  Narayanan}]{ma2025multilabel}
Ma, M.; Chochlakis, G.; Pandiyan, N.~M.; Thomason, J.; and Narayanan, S. 2025.
\newblock Large Language Models Do Multi-Label Classification Differently.
\newblock In \emph{Proceedings of the 2025 Conference on Empirical Methods in
  Natural Language Processing}, 2472--2495. Association for Computational
  Linguistics.

\bibitem[{Manakul, Liusie, and Gales(2023)}]{manakul2023selfcheckgpt}
Manakul, P.; Liusie, A.; and Gales, M. J.~F. 2023.
\newblock {SelfCheckGPT}: Zero-Resource Black-Box Hallucination Detection for
  Generative Large Language Models.
\newblock In \emph{Proceedings of the 2023 Conference on Empirical Methods in
  Natural Language Processing}, 9004--9017. Association for Computational
  Linguistics.

\bibitem[{Mukku et~al.(2023)Mukku, Soni, Aggarwal, Rana, Yenigalla, Patange,
  and Mohan}]{mukku2023insightnet}
Mukku, S.~S.; Soni, M.; Aggarwal, C.; Rana, J.; Yenigalla, P.; Patange, R.; and
  Mohan, S. 2023.
\newblock {InsightNet}: Structured Insight Mining from Customer Feedback.
\newblock In \emph{Proceedings of the 2023 Conference on Empirical Methods in
  Natural Language Processing: Industry Track}, 552--566. Association for
  Computational Linguistics.

\bibitem[{Ouyang et~al.(2025)Ouyang, Zhang, Harman, and
  Wang}]{ouyang2025nondeterminism}
Ouyang, S.; Zhang, J.~M.; Harman, M.; and Wang, M. 2025.
\newblock An Empirical Study of the Non-Determinism of {ChatGPT} in Code
  Generation.
\newblock \emph{ACM Transactions on Software Engineering and Methodology},
  34(2): 1--28.

\bibitem[{Pham et~al.(2024)Pham, Hoyle, Sun, Resnik, and
  Iyyer}]{pham2024topicgpt}
Pham, C.~M.; Hoyle, A.; Sun, S.; Resnik, P.; and Iyyer, M. 2024.
\newblock {TopicGPT}: A Prompt-Based Topic Modeling Framework.
\newblock In \emph{Proceedings of the 2024 Conference of the North American
  Chapter of the Association for Computational Linguistics: Human Language
  Technologies}, 2956--2984. Association for Computational Linguistics.

\bibitem[{Reimers and Gurevych(2019)}]{reimers2019sbert}
Reimers, N.; and Gurevych, I. 2019.
\newblock Sentence-{BERT}: Sentence Embeddings Using Siamese {BERT}-Networks.
\newblock In \emph{Proceedings of the 2019 Conference on Empirical Methods in
  Natural Language Processing and the 9th International Joint Conference on
  Natural Language Processing}, 3982--3992. Association for Computational
  Linguistics.

\bibitem[{Sclar et~al.(2024)Sclar, Choi, Tsvetkov, and
  Suhr}]{sclar2024quantifying}
Sclar, M.; Choi, Y.; Tsvetkov, Y.; and Suhr, A. 2024.
\newblock Quantifying Language Models' Sensitivity to Spurious Features in
  Prompt Design or: How I Learned to Start Worrying about Prompt Formatting.
\newblock In \emph{International Conference on Learning Representations}.

\bibitem[{Sedova et~al.(2024)Sedova, Litschko, Frassinelli, Roth, and
  Plank}]{sedova2024know}
Sedova, A.; Litschko, R.; Frassinelli, D.; Roth, B.; and Plank, B. 2024.
\newblock To Know or Not to Know? Analyzing Self-Consistency of Large Language
  Models under Ambiguity.
\newblock In \emph{Findings of the Association for Computational Linguistics:
  EMNLP 2024}, 17203--17217. Association for Computational Linguistics.

\bibitem[{Song et~al.(2025)Song, Wang, Li, and Lin}]{song2025good}
Song, Y.; Wang, G.; Li, S.; and Lin, B.~Y. 2025.
\newblock The Good, The Bad, and The Greedy: Evaluation of {LLM}s Should Not
  Ignore Non-Determinism.
\newblock In \emph{Proceedings of the 2025 Conference of the Nations of the
  Americas Chapter of the Association for Computational Linguistics: Human
  Language Technologies}, 4195--4206. Association for Computational
  Linguistics.

\bibitem[{Tabatabaei et~al.(2025)Tabatabaei, Fancher, Parsons, and
  Askari}]{tabatabaei2025hierarchical}
Tabatabaei, S.~A.; Fancher, S.; Parsons, M.; and Askari, A. 2025.
\newblock Can Large Language Models Serve as Effective Classifiers for
  Hierarchical Multi-Label Classification of Scientific Documents at Industrial
  Scale?
\newblock In \emph{Proceedings of the 31st International Conference on
  Computational Linguistics: Industry Track}, 163--174. Association for
  Computational Linguistics.

\bibitem[{Wang et~al.(2023)Wang, Wei, Schuurmans, Le, Chi, Narang, Chowdhery,
  and Zhou}]{wang2023selfconsistency}
Wang, X.; Wei, J.; Schuurmans, D.; Le, Q.~V.; Chi, E.~H.; Narang, S.;
  Chowdhery, A.; and Zhou, D. 2023.
\newblock Self-Consistency Improves Chain of Thought Reasoning in Language
  Models.
\newblock In \emph{International Conference on Learning Representations}.

\bibitem[{Xiong et~al.(2024)Xiong, Hu, Lu, Li, Fu, He, and
  Hooi}]{xiong2024uncertainty}
Xiong, M.; Hu, Z.; Lu, X.; Li, Y.; Fu, J.; He, J.; and Hooi, B. 2024.
\newblock Can {LLM}s Express Their Uncertainty? An Empirical Evaluation of
  Confidence Elicitation in {LLM}s.
\newblock In \emph{International Conference on Learning Representations}.

\bibitem[{Yao et~al.(2024)Yao, Shinn, Razavi, and Narasimhan}]{yao2024taubench}
Yao, S.; Shinn, N.; Razavi, P.; and Narasimhan, K. 2024.
\newblock $\tau$-bench: A Benchmark for Tool-Agent-User Interaction in
  Real-World Domains.
\newblock arXiv:2406.12045.

\end{thebibliography}

\end{document}